\documentclass[final, 12pt]{clear2023} 
\graphicspath{ {./images/} }
\usepackage{times}
\usepackage{appendix}
\usepackage{chngcntr}

\title[SCARY Dataset]{The Structurally Complex with Additive Parent Causality (SCARY) Dataset}

\author{
 \Name{Jarry Chen}
 \Email{jarry.chen@ibm.com}
 \AND
 \Name{Haytham M.Fayek}
 \Email{haytham.fayek@ieee.org} \\
 \addr RMIT University}
 
\makeatletter
\def\@evenhead{\hfil\shortauthors\hfil}
\def\shortauthors{CHEN AND FAYEK}
\makeatother

\begin{document}
\maketitle

\begin{abstract}
Causal datasets play a critical role in advancing the field of causality. However, existing datasets often lack the complexity of real-world issues such as selection bias, unfaithful data, and confounding. To address this gap, we propose a new synthetic causal dataset, the Structurally Complex with Additive paRent causalitY (SCARY) dataset, which includes the following features. The dataset comprises 40 scenarios, each generated with three different seeds, allowing researchers to leverage relevant subsets of the dataset. Additionally, we use two different data generation mechanisms for generating the causal relationship between parents and child nodes, including linear and mixed causal mechanisms with multiple sub-types.
Our dataset generator is inspired by the Causal Discovery Toolbox and generates only additive models. The dataset has a Varsortability \citep{reisach2021beware} of 0.5. Our SCARY dataset provides a valuable resource for researchers to explore causal discovery under more realistic scenarios. The dataset is available at https://github.com/JayJayc/SCARY.
\end{abstract}


\section{Introduction}

Causal discovery is a fundamental task in many scientific fields, including medicine, psychology, and economics. In recent years, there has been a growing interest in developing and evaluating algorithms for causal discovery using datasets with varying degrees of complexity and causality. While many datasets include confounding, few include unfaithfulness and selection bias, which are data issues or assumptions used for causal discovery.

In this paper, we present the Structurally Complex with Additive paRent causalitY (SCARY) dataset, a novel dataset designed to evaluate the performance of causal discovery algorithms under different scenarios. The SCARY dataset includes three forms of data issues: unfaithfulness, causal sufficiency, and selection bias. We also provide a detailed description of our data generator, which employs various causal mechanisms to simulate the effects of different types of causal relationships.

By incorporating these forms of data issues and generating data using different causal mechanisms, the SCARY dataset offers a unique opportunity to explore the strengths and limitations of different causal discovery algorithms under varying degrees of complexity and causality. Ultimately, the SCARY dataset aims to advance the field of causal discovery by providing a more realistic and challenging test bed for evaluating and comparing causal discovery methods.

\newpage

\begin{figure}[t]
\centering
\includegraphics[scale=0.45]{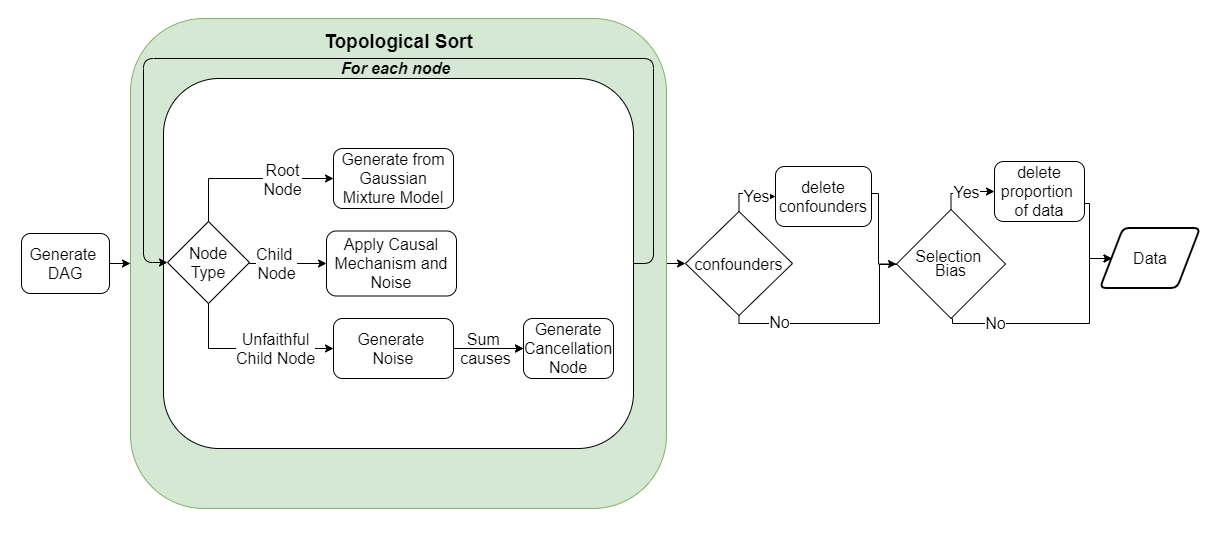}
\caption{Data generation process}\label{fig:data gen}
\end{figure}

\section{Data Generator}
\label{sec:headings}

Our data generator's unique algorithm (see Figure \ref{fig:data gen}) ensures that the generated DAGs are more problematic which is a better representation of real-world datasets, with varying degrees of complexity and causality. The root nodes' data are generated using a Gaussian mixture model, providing a basis for simulating the initial variables' values in the system. The use of a spherical covariance type ensures that the variance of the generated data is constant across all dimensions, thereby preventing bias towards any particular feature.

\begin{figure}[t]
\centering
\includegraphics[scale=0.5]{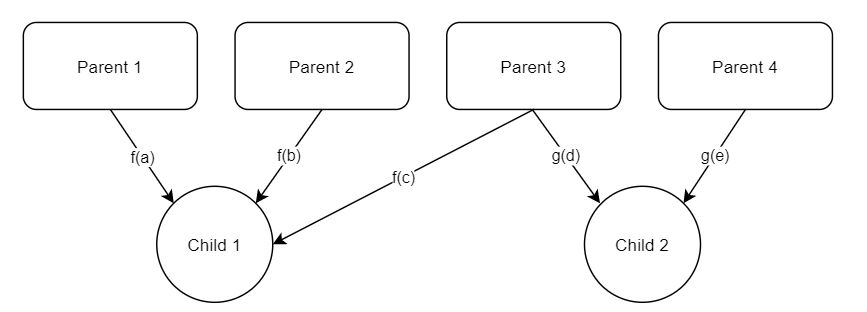}
\caption{Causal mechanism}\label{fig:causal mech}
\end{figure}

Generating data for the child nodes is a more complex process that involves applying the selected causal mechanisms to their parents' data (see Figure \ref{fig:causal mech}). We use various causal mechanisms, including linear, non-linear (e.g., polynomial and sigmoid), and mixed functions, to simulate the effects of different types of causal relationships. To add some variability to the data, we also incorporate a mixed mechanism option that selects a random mix of mechanisms for each child's causal relationship with its parents. This approach introduces additional complexity to the dataset, which can challenge the assumptions of causal discovery algorithms.

Therefore, our generated dataset is an ideal test bed for evaluating and comparing causal discovery methods and their ability to handle catastrophic failure in their assumptions. It provides a unique opportunity to explore the strengths and limitations of different causal discovery algorithms under varying degrees of complexity and causality.

\subsection{Unfaithfulness}

For unfaithfulness, we use near-failures of faithfulness rather than failures of faithfulness to generate data. Due to practical limitations in data collection and selection bias, the use of near-failures of faithfulness in data generation is more appropriate for evaluating causal discovery methods, as it is more likely that algorithms will encounter near-failures of faithfulness than exact failures. The measure zero argument \citep{spirtes2000causation} claims that the set of parameter values that cancel exactly along a path is infinitesimally small, and the probability of two paths canceling exactly is zero. However, the unlikely stability of these parameters in real-world data \citep{pearl2009causality} makes using near-failures of faithfulness more practical and relevant for evaluating causal discovery methods.

\subsection{Causal Sufficiency}

For causal sufficiency, we assume that it is unlikely that all relevant confounders are observed and data has been collected. This is often the case in observational data, and it is rarely satisfied. Our data generation process does not guarantee a non-arid, arid, bow or bow-free graph, and therefore should only be used to study the impacts of unobserved confounders on Bayesian Network structure learning algorithms as opposed to Acyclic Directed Mixed Graphs (ADMG) which assumes the graphs are arid and/or bow-free.

\subsection{Selection Bias}

Selection bias occurs when observations are excluded from a sample, making it not representative of the population or causal data generation process. Excluding a sub-population can be represented as an unobserved common response variable and will not be generated by our data generation process. We also avoid introducing unfaithfulness through selection bias, by not using full canceling selection bias, therefore circumventing the issue where only the sub-population that has canceling paths are in the sample but the same cancellation will not be observed on average across the whole population \citep{weinberger2018faithfulness}. Selection bias differs from confounding bias as it only affects the inclusion of a data point in the sample, leading to under- or over-representation of certain sub-populations and incorrect conclusions about causal relationships. Therefore, it is crucial to consider selection bias in the data generation process for causal discovery studies.

\section{The SCARY Dataset}\label{sec:headings}

\begin{figure}[t]
\centering
\includegraphics[scale=0.5]{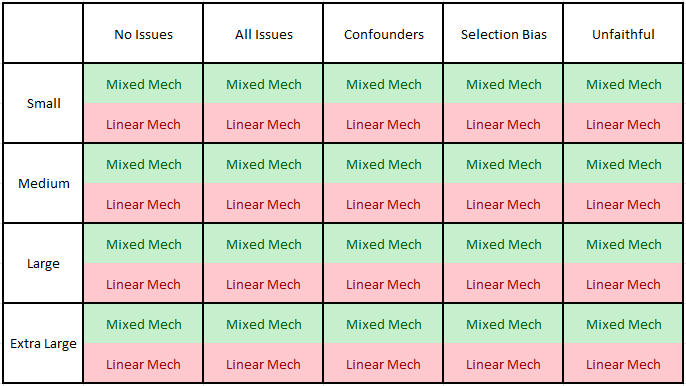}
\caption{Data generation matrix}\label{fig:scenario matrix}
\end{figure}

The SCARY dataset is composed of 240 sub-datasets with two density levels, each containing 2500 samples, and 40 unique generator configurations (see Figure \ref{fig:scenario matrix}), each of which employs 3 distinct seeds. To ensure a comprehensive evaluation of the impact of each issue type on the overall dataset, we have created a diverse set of scenarios that combine various DAG sizes and issue types. The dataset includes scenarios with DAG sizes of small (10 nodes), medium (15 nodes), large (25 nodes), and extra-large (50 nodes), with a proportional scaling of issues at a 10:1 ratio, always rounded up. For instance, the 15 node scenario incorporates 2 issues, while the 25 node scenario incorporates 3 issues to maintain the same scaling ratio. We have also included DAGs in the dataset that do not exhibit any issues, which can be used as a benchmark.

\subsection{Causal Mechanism}

The dataset is divided into two parts: one with data generated through a linear mechanism and the other with data generated through mixed mechanisms. In the case of the mixed mechanisms, a file is included that specifies the parent-to-child mechanism used to generate the data, allowing researchers to trace the origins of the data. The use of mixed mechanisms ensures that algorithms cannot make assumptions about relationships between nodes based on just one parent-child relationship, which better reflects the diversity of causal functions that can occur in the real world.

\subsection{Varosortability}

The dataset's Varsortability \citep{reisach2021beware} is approximately 0.5, which indicates a lack of agreement between the partial order induced by the marginal variances and all pathwise descendant relations implied by the causal structure. The additive causal relationship between each child and its parents allows us to attain this Varsortability value. Nonetheless, this approach is not ideal for causal inference studies and should only be used for causal discovery.

\section{Conclusion}
\label{sec:headings}
The paper presents a novel synthetic causal dataset, the Structurally Complex with Additive paRent causalitY (SCARY) dataset, to address the limitations of existing datasets that often lack the complexity of real-world issues such as selection bias, unfaithful data, and confounding. The SCARY dataset includes 40 scenarios, each generated with three different seeds, using two different data generation mechanisms for generating the causal relationship between parents and child nodes. The dataset offers a unique opportunity to explore the strengths and limitations of different causal discovery algorithms under varying degrees of complexity and causality, thereby advancing the field of causal discovery. The paper concludes that the generated dataset is an ideal test bed for evaluating and comparing causal discovery methods and their ability to handle catastrophic failure in their assumptions. The dataset is available at https://github.com/JayJayc/SCARY.

\newpage
\appendix
\counterwithin{figure}{section}

\renewcommand{\thefigure}{\Alph{figure}}

\section{\\Sample DAGs}
\begin{figure}[h!]
\centering
\includegraphics[scale=0.365]{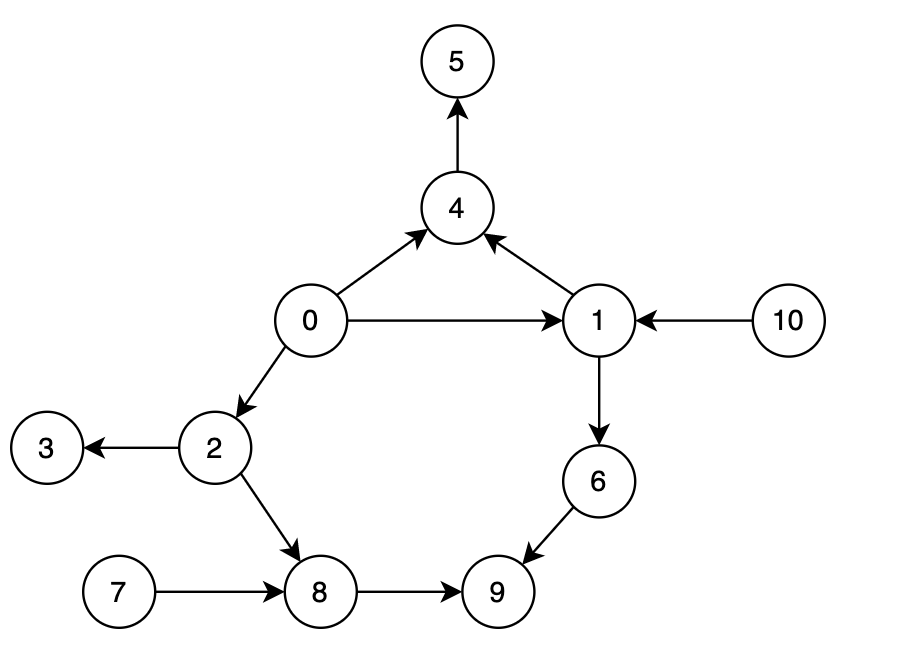}
\caption{Sample DAG with confounder}
\label{fig:confounder_plot}
\end{figure}

\begin{figure}[h!]
\centering
\includegraphics[scale=0.24]{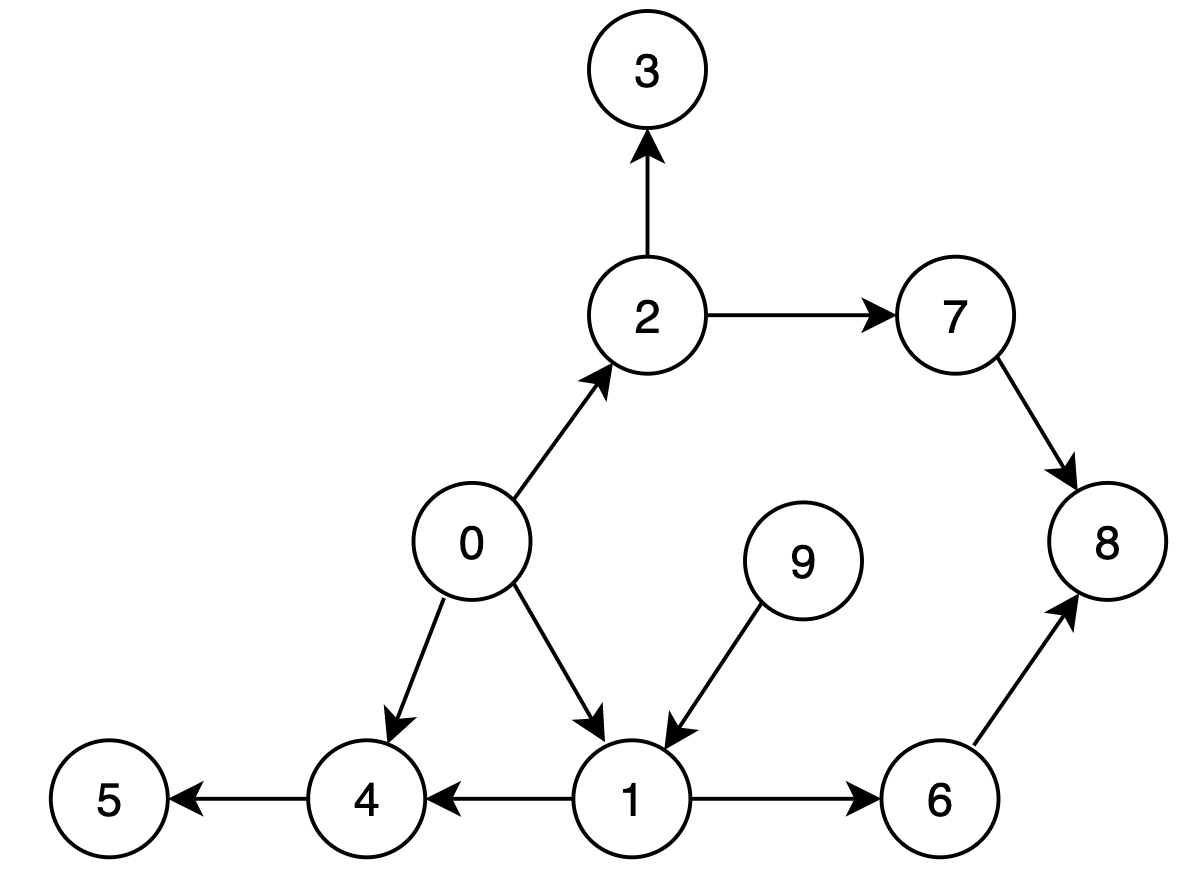}
\caption{Sample DAG without confounder}
\label{fig:plot}
\end{figure}

Figure \ref{fig:confounder_plot} shows a sample of a sparse 10-node directed acyclic graph (DAG) that includes all issues, with an extra confounder node. Each node represents a variable, and each edge represents a causal relationship between two variables. The DAG includes a total of 10 nodes and one confounder node, resulting in 11 nodes in total. Figure \ref{fig:plot} shows the same sparse 10-node directed acyclic graph (DAG) as before, but with the confounder node removed.s

\bibliography{references}

\begin{thebibliography}{4}
\providecommand{\natexlab}[1]{#1}
\providecommand{\url}[1]{\texttt{#1}}
\expandafter\ifx\csname urlstyle\endcsname\relax
  \providecommand{\doi}[1]{doi: #1}\else
  \providecommand{\doi}{doi: \begingroup \urlstyle{rm}\Url}\fi

\bibitem[Pearl(2009)]{pearl2009causality}
Judea Pearl.
\newblock \emph{Causality}.
\newblock Cambridge university press, 2009.

\bibitem[Reisach et~al.(2021)Reisach, Seiler, and Weichwald]{reisach2021beware}
Alexander Reisach, Christof Seiler, and Sebastian Weichwald.
\newblock Beware of the simulated dag! causal discovery benchmarks may be easy
  to game.
\newblock \emph{Advances in Neural Information Processing Systems},
  34:\penalty0 27772--27784, 2021.

\bibitem[Spirtes et~al.(2000)Spirtes, Glymour, Scheines, and
  Heckerman]{spirtes2000causation}
Peter Spirtes, Clark~N Glymour, Richard Scheines, and David Heckerman.
\newblock \emph{Causation, prediction, and search}.
\newblock MIT press, 2000.

\bibitem[Weinberger(2018)]{weinberger2018faithfulness}
Naftali Weinberger.
\newblock Faithfulness, coordination and causal coincidences.
\newblock \emph{Erkenntnis}, 83\penalty0 (2):\penalty0 113--133, 2018.

\end{thebibliography}
\end{document}